\documentclass[11pt]{article}

\usepackage[preprint]{acl}

\usepackage{times}
\usepackage{latexsym}
\usepackage{subcaption}
\usepackage{amsmath,amsfonts,amssymb}
\usepackage[T1]{fontenc}
\usepackage{paralist}
\usepackage[utf8]{inputenc}

\usepackage{microtype}

\usepackage{inconsolata}

\usepackage{graphicx}
\usepackage{booktabs}
\usepackage{url}
\usepackage{hyperref}

\usepackage{annotates}

\newcommand{\new}[1]{\textcolor{blue} {#1}}
\renewcommand{\new}[1]{#1}

\title{Off-the-Shelf Language-Vision Encoders Predict \\ Human Gaze Preferences
 in Visual World Experiments}

\author{Rahul Murali Shankar \\
  IMS \\
  University of Stuttgart \\
  \And
  Titus von der Malsburg \\
  Institute of Linguistics \\
  University of Stuttgart \\
  \texttt{\{st196277@stud,titus.von-der-malsburg@ling,pado@ims\}.uni-stuttgart.de} \\\And
  Sebastian Padó \\
  IMS \\
  University of Stuttgart \\
}

\begin{document}
\maketitle
\begin{abstract}
\new{A central question for computational psycholinguistics is to what extent neural language models can serve as cognitively adequate  models of human language processing.} However, work has been overwhelmingly focused on the unimodal case of written or spoken language. In contrast, multimodal paradigms, like visual world experiments that present participants with both visual and linguistic input simultaneously, have been neglected. In this paper, we present a novel approach that predicts \new{patterns of gaze preferences in visual world studies. To do so, it derives target preferences by applying an attribution method to a simple bi-modal CLIP bi-encoder.} We demonstrate the ability of this approach to \new{qualitatively} replicate the gaze patterns found in a seminal English visual world study demonstrating human predictive processing. It does so without a generative architecture and without the need for fine-tuning, despite not being trained for the task.
\end{abstract}

\begin{figure*}[b!th]
    \centering
    \begin{subfigure}[t]{0.235\linewidth}
    \centering
        \includegraphics[width=\linewidth]{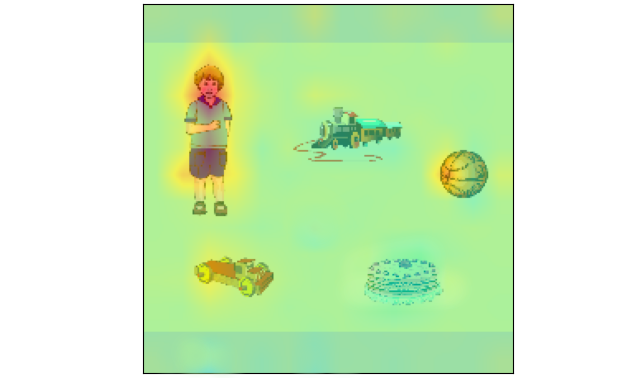}
        \subcaption{'the boy'}
    \end{subfigure}\hfill
    \begin{subfigure}[t]{0.235\linewidth}
        \centering
        \includegraphics[width=\linewidth]{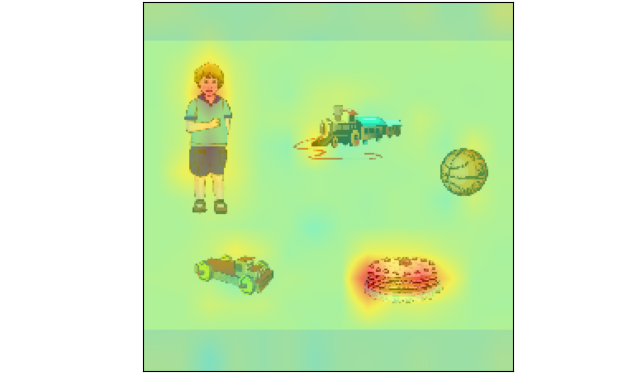}
        \subcaption{'the boy will eat'}
    \end{subfigure}\hfill
    \begin{subfigure}[t]{0.235\linewidth}
        \centering
        \includegraphics[width=\linewidth]{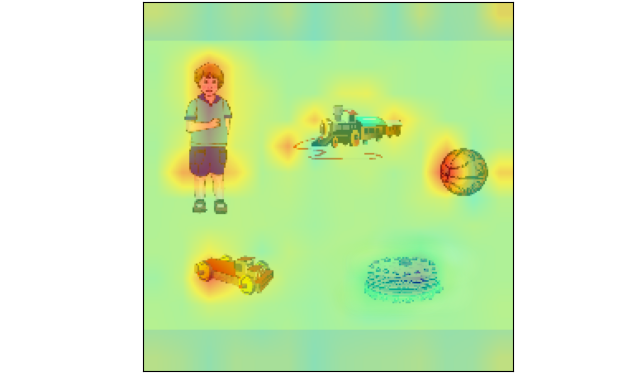}
        \subcaption{'the boy will move'}
    \end{subfigure}\hfill
    \begin{subfigure}[t]{0.285\linewidth}
        \centering
        \includegraphics[width=0.825\linewidth]{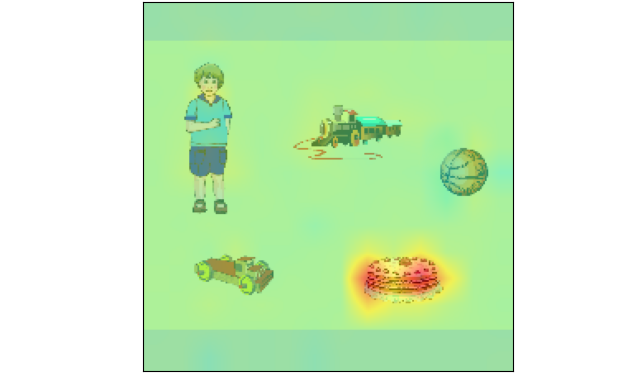}
        \subcaption{'the boy will eat/move the cake'}
    \end{subfigure}
    \caption{\new{CLIP attributions (heatmaps)  illustrate human gaze patterns found by \citet{altmann99}.}}
    \label{fig:heatmaps}
\end{figure*}

\section{Introduction}

Neural networks, in particular language models (LMs), dominate both applied and theoretical approaches to modeling linguistic phenomena \citep{LinzenBaroni2021}.
%Specifically, \textit{transformers} \citep{vaswani2017attention} have emerged as a major success story, supplanting an earlier generation of recurrent network architectures by offering improved scalability and parallelization, and, on many tasks, greater parameter efficiency (but see \citealt{BeckEtAl2024}).
% Virtually all state-of-the-art large language models (LLMs) are now based on transformers \citep{NEURIPS2020_1457c0d6}.
From a psycholinguistic point of view, the success of neural LMs stands in stark contrast to their simplistic next-word prediction objective.
This raises the question: To what extent can neural LMs account for the cognitive processes underlying human language processing? Over the last years, neural LMs have been able to account for a wealth of findings in human language comprehension and production 
\citep[e.g.,][]{Wilcox2020OnTPA,frank-2024-neural}.
%, including patterns in reaction times \citep{Wilcox2020OnTPA}, event-related potentials \citep{frank-2024-neural}, and functional MRI activity \citep{Toneva2019InterpretingAIA} during sentence comprehension.
and expectations about their capabilities in cognitive modeling have risen \citep{CuskleyEtAl2024,https://doi.org/10.1111/lnc3.70001}.

In this paper, we consider an experimental paradigm, underexplored in the CL community, that combines linguistic and visual processing: the \textit{visual-world paradigm}  \citep[see][for a recent overview]{Ito_Knoeferle_2023}. Visual world studies record and analyze participants' eye movements across a real scene or a computer display while the participants listen to linguistic stimuli or sounds. By  observing eye movements during the presentation of the linguistic input, researchers can test  rich hypotheses about incremental processing of multimodal input \citep{magnuson2019fixations}.

Figure \ref{fig:heatmaps} illustrates how visual world studies can demonstrate human predictive processing \citep{altmann99}. The heatmaps are not human data but attributions computed by a CLIP model (yellow/red: high attribution; see Section~\ref{sec:overview}). They mirror the gaze pattern that \citet{altmann99} observed in human participants:
 In (\ref{fig:heatmaps}a), after the input ``\textit{the boy}'', participants predominantly look at the boy, establishing the referent for the subject. When the input continues with ``\dots \textit{will eat}'', shown in (\ref{fig:heatmaps}b), the locus of preferred fixations switches to the cake, although it has not been mentioned verbally, indicating that participants anticipate likely objects of the eating event. Similarly, the alternative continuation in (\ref{fig:heatmaps}c), ``\dots \textit{will move}'' makes participants look at objects that are easily movable, like toy trains and balls. For the completed sentence in (\ref{fig:heatmaps}d), fixations converge on the sentence-final object (``the cake'').

\begin{figure}[tb!]
\centering
\includegraphics[width=\linewidth]{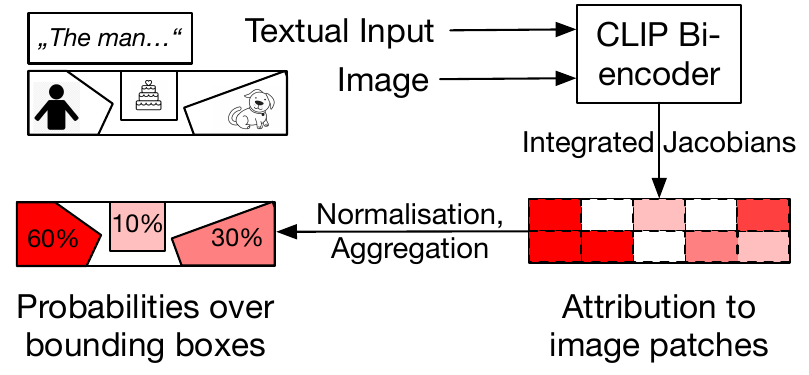}
\caption{Structure of our approach}
\vspace*{-1.5em}
\label{fig:model_structure}
\end{figure}

In contrast to (unimodal) language processing, there is little recent work on the modeling of visual world studies, despite the emergence of vision-language models (VLMs) which capture language-vision interactions. This is presumably because models such as Llava \citep{Liu2023VisualInstruction} and Qwen-VL \citep{Bai2025Qwen25VL} can generate either text or images, but it is not straightforward --  a research question in itself --  how to use such models to generate \new{gaze patterns} for text-image combinations.

In this paper, we propose an approach, shown in Figure\ \ref{fig:model_structure}. We use one of the simplest, pre-generative, language-vision architectures, the contrastively tuned bi-encoder CLIP model \citep{clip}, which computes a similarity between a visual and a linguistic input. We then apply an explainability method, Integrated Jacobians \citep{moeller-etal-2024-approximate}, which computes \textit{attribution scores} that quantify the contribution of each pair of linguistic token and image patch. We show that these attribution scores -- suitably normalized and aggregated -- \new{predict human gaze preferences.}
%are directly linearly related to human %fixations:
\new{As a proof of concept, our approach qualitatively replicates} the effects found in a seminal English visual world study \citep{altmann99}.\footnote{We use \textit{approach} to refer to our proposed method to distinguish it terminologically from other uses of 'model', notably the CLIP models and the statistical analysis models.}

This is notable for three reasons: (a), our approach does not require any \new{fine-tuning: predictions for human gaze preferences are derived from pre-trained CLIP models, given choices of model, normalization, and aggregation}; (b), our approach \new{only requires the annotation of bounding boxes for the objects in the images but no further labeling}; (c), the approach accounts for predictive behavior in human language processing, such as the \new{preference for} the cake in Figure (\ref{fig:heatmaps}b) \new{after} the verb 'eat', although CLIP is an encoder-only, non-generative model. \new{This suggests that the effect can arise from correlational structure in pre-training data.}

\section{Background and Related Work}

\paragraph{The Visual World Paradigm.}
The visual world paradigm provides a fine-grained, continuous-time measure of how listeners integrate unfolding speech with concurrent visual information.  A hallmark finding of this research is that human language comprehension is highly incremental and predictive; for example, listeners use verb selectional restrictions to proactively fixate on plausible visual referents before the target noun is spoken \citep{altmann99}.  Beyond predictive processing, the paradigm has been instrumental in illuminating a wide range of linguistic phenomena.  These include tracking phonological competition during spoken word recognition \citep{allopenna1998tracking}, demonstrating the immediate use of visual context to resolve syntactic ambiguities \citep{tanenhaus1995}, and tracing the rapid computation of pragmatic inferences \citep{sedivy1999}.  Together, these gaze patterns demonstrate that linguistic input rapidly guides the allocation of visual attention by continuously updating multimodal event representations in working memory \citep{huettig2011using}.  However, because the concurrent visual display can pre-activate lexical candidates and alter competition dynamics, \citet{huettig2011using} caution that these findings should not be generalized to language processing in the absence of visual input.

\paragraph{Computational Models of Human Language Processing.} Neural networks have been proposed early on as models of human language processing \citep{Rumelhart1986PastTense}. The neural network-related developments of the last decade have led to models that can account for, e.g., patterns in reaction times \citep{Wilcox2020OnTPA}, event-related potentials \citep{frank-2024-neural}, and functional MRI activity during sentence comprehension \citep{Toneva2019InterpretingAIA}. However, the picture is far from clear, with models underpredicting processing
difficulty for garden-path sentences \citep{VanSchijndelLinzen2021}, and concerns about the impact of model size \citep{OhSchuler2023}.

\paragraph{Models of Visual World Experiments.} To our knowledge, the only study that explicitly models stimulus-level \new{gaze} patterns for visual world experiments with broad-coverage models of language is \citet{allopenna1998tracking}. They query the TRACE model of speech perception \citep{mcclelland1986trace} to compute activations for various words on a fine-grained time scale. However, their model does not establish a cross-modal mapping: it works only on the linguistic input and its predictions are mapped onto visual referents by way of linguistic labels ("boy"). Thus, the cross-modal link is still specified manually, and the model cannot account for properties of the visual stimuli such as their visual saliency \citep{BRUCE201595}. Other analyses are concerned with integrating theories of eye movement with results from visual world experiments \citep{mirman2008statistical}, which we are not concerned with here.

%\paragraph{Eye-Tracking in NLP} 
%\citet{10.1145/3649902.3656356}
%\citet{sauberli-etal-2026-controlling}
%\citet{barrett-etal-2018-sequence}

%\titus{Do we need this?  I feel this is only tangentially related and the paper would perhaps benefit more from some background on CLIP models?  I’m unsure.}

\section{Method}

\subsection{Overview}
\label{sec:overview}
Figure \ref{fig:model_structure} visualizes our approach. For a pair of a language input and an image, our approach produces a \new{distribution of attributions over image patches, given the language input}. This prediction process does not require bounding boxes. However, since \new{visual world studies typically analyze gazes at the level of objects in a scene}, the approach also includes a step to aggregate the \new{attributions} at the level of objects as described by bounding boxes.

 This section outlines the scope and linking hypothesis of our approach (§3.2) before describing the three processing steps in detail: encoding,  attribution, and normalization/aggregation (§3.3-3.5).
% We close with a discussion of modes of presenting the experimental stimuli (§3.6).
 We  submit an archive of our implementation as supplementary material (CC license).

\subsection{Scope and Linking Hypothesis}

The first question that needs to be addressed is why we should believe that attributions from CLIP models should \new{be related to human gaze preferences}.

On the experimental side, it is generally assumed that eye movements during the presentation of linguistic input are involved in establishing associations between the visual field and the linguistic input \citep{cooper1974control}. There is considerable debate about the details of this process \citep{degen2021seeing} and the influence of other cognitive processes \citep{wolfe2017five} and prior context \citep{yoon2018influence}. However, for the simple, static visual materials and the token-level predictions we consider here (cf. Section~\ref{sec:overview}), it is arguably sufficient to assume that \new{gaze preferences} reflect the processing of the linguistic input, as assumed previously by \citet{allopenna1998tracking}.

On the model side, \citet{moeller2025explainingvisionlanguagesimilaritiesdual} found that CLIP models learn to categorize objects in their visual input, despite a lack of explicit training for this task, since the resulting knowledge is helpful for the objective of distinguishing between relevant and irrelevant captions. Consequently, when presented with visual world materials, where there is, by design, a good correspondence between the linguistic input and (a subset of) the shown objects, we generally expect positive visual attributions for relevant objects. This constitutes a good match with the human predictions outlined above.
%and provides a measurement of how well the model manages to identify 'relevant' objects. 

Note that in our approach, the granularity of the predictions is more coarse-grained than in human experiments: we do not predict \new{gaze preferences} at specific time points, but for each input token (according to the encoder's tokenization). Analysis of visual world data in psycholinguistics is generally carried out at the level of time intervals \citep{Ito_Knoeferle_2023}, but these are almost always anchored in terms of input words. Our approach is therefore sufficiently fine-grained to account for many visual world effects, unless the research focus is on exact time courses \citep{mirman2008statistical}.

\subsection{Step 1: CLIP Bi-encoders}

In the era of LLMs, encoder models that map inputs into high-dimensional representations
(``embeddings'') continue to play an important role in NLP.
%, in particular in applications that use textual similarity. 
%These range from semantic search \cite{alatrash2024100193} to NLG evaluation \cite{chollampatt-etal-2025-cross}, graph reasoning \cite{plenz-etal-2023-similarity} and RAG \cite{neurips2020_6b493230_rag}. 
A  popular encoder family are \textit{dual-encoder models} such as SBERT \cite{reimers-gurevych-2019-sentence} which encode both input texts independently of one another and then compute similarity as a function of the two representations.

Bi-Encoders have also been applied successfully to model language--image similarity, arguably most successfully by the Contrastive Language-Image Pre-Training (CLIP) approach \citep{clip}. \new{CLIP jointly pre-trains text and  image encoders on the task of pairing images with captions on a very large dataset.  CLIP  is compatible with different encoder choices, but standard choices are (a) Vision Transformers of various types for the image encoder, and (b) Transformer-based language models for the text encoder.} 
Despite the simplicity of their training objective, CLIP representations are robust across domains and tasks and highly informative for downstream applications, such as image classification \citep{medclip} or visual question answering \citep{shen2021clipbenefitvisionandlanguagetasks}. 

From a psycholinguistic modeling point of view, CLIP models have an additional advantages over generative VLMs in that encoder architectures are arguably a better fit as models of perception than decoder architectures optimized for generation.

%; (b) they are 
%conceptually simpler and practically smaller.
%, requiring less memory and computation.

  %  \rahul{This note is for Table 1. See \citet{clip} Table 10 (p. 40)/Table 11 (p. 43), \citet{cherti2023reproducible} Table 1 (p. 2)/Table 5/6/7 (p. 28), the Spearman rho may change due to this}

\begin{table*}[tb!h]
    \centering
    \begin{tabular}{lllc}
    \toprule
       Model  & Params. & Trained on & ImageNet ZS Acc. \\
       \midrule
1 OpenAI ViT-B/16 & 150M & WIT-400M & 68.6\% \\
2 OpenCLIP ViT-B/16  & 150M & datacomp\_xl\_s13b\_b90k & 73.5\% \\
3 OpenCLIP ViT-L/14 & 427M & datacomp\_xl\_s13b\_b90k & 79.2\% \\
4 OpenCLIP ViT-L/14 (quickgelu) & 427M & dfn2b & 81.4\% \\
         \bottomrule
    \end{tabular}
    \caption{The CLIP models we consider. ImageNet ZS Acc.: ImageNet zero-shot accuracy \\ (Sources: \citet{clip} for 1; \citet{gadre2023datacompsearchgenerationmultimodal} for 2 and 3; \citet{fang2023datafilteringnetworks} for 4)}
    \label{tab:models}
\end{table*}

\subsection{Step 2: Attribution}

CLIP models are designed to merely produce 'black-box' cross-modal similarities between their inputs based on their embeddings. This is clearly not enough for predictions about human perceptions. This information can be provided by methods from the explainability domain. Specifically, 
\textit{local feature attribution methods} explain a given prediction by assigning scores to individual input features \citep{murdoch}.

A prominent such method for encoders is Integrated Gradients (IG, \citealt{ig}). It 
attributes a scalar model prediction $f(s)$ back onto individual input features $i$.
%
%by computing the gradient of $f$ on a number $N$ %of interpolations between the actual input $s$ %and an uninformative reference input $r$ and %summing up. Define this interpolation as %$x(\alpha) = r + \alpha (s-r)$. Then: \vspace*%{-1em}
%
%\begin{small}
%  \begin{align*}
%f(s)  \approx \sum_i (s-r)_i G_i  \text{ where }
%G_i   \approx  \frac{1}{N} \sum_{n=1}^N %\frac{\partial f(x(\alpha_n))}{\partial x_i}
%  \end{align*}
%  \end{small}
% IG has a number of desirable properties, among others that the decomposition is provably correct. 
%
%Since IG returns a \textit{vector} $G_i$ of %feature attributions $i$, it is
IG is however
not  applicable to dual encoders, since the similarities arise from the \textit{interactions} between features of the two inputs rather than from features of a single input.  \citet{moeller-etal-2024-approximate} introduce Integrated Jacobians (IJ), a generalization of IG to dual encoders $f(s, t)$ whose similarity function is defined as an inner product over two independent, vector-valued encoders, such as CLIP.
%: $f(s, t) = g(s)^{\intercal} h(t)$.
In this case, the function $f$ can be decomposed into individual contributions of interactions between input features of $s$ and $t$: 
\begin{align*}
f(s,t)  \approx & \sum_{ij} (s - r_s)_i J^s_{ik} J^t_{kj} (t-r_t)_j
%\text{with} \\
% (J^s)_{ki}  \approx & \frac{1}{N} \sum_{n=1}^N \frac{\partial g_k(x(\alpha_n))}{\partial x_i}  
\end{align*}
\new{Here, $r_s$ and $r_t$ are uninformative reference inputs for $s$ and $t$, respectively.\footnote{\new{\citet{moeller-etal-2024-approximate} establish that approximate attributions, which assume that $f(r) \approx 0$, agree very well with exact attributions, so that $f(r)$ can be ignored in practice.}} $J^s$ is the Jacobian of the encoder for input $s$, i.e., the matrix of partial derivations of all output dimensions w.r.t. all input dimensions.
%evaluated on $N$ interpolations $\alpha$ between $r$ and $s$. $J^t$ is defined in parallel for $t$. 
When applied to the CLIP case, the $J^s$ and $J^t$   matrices provides one score for each pairs of tokens and image patches, describing which parts of the linguistic input and the visual input do or do not match one another \citep{moeller2025explainingvisionlanguagesimilaritiesdual}.}

\subsection{Step 3: Normalization of Attributions}
\label{sec:normalization}

By definition of Integrated Jacobians, the scores in the attribution
matrix sum to the similarity between language and image. The
contribution of individual pairs can however also be negative, viz.\
in cases of strong mismatches (cf. Figure
\ref{fig:heatmaps}(c)). Consequently, the 'raw' attributions \new{do not
form a probability distribution over the image.}

\new{Since negative attributions are generally considered to be unreliable
  \citep{moeller-etal-2024-approximate}, we apply ReLU normalization. It discards negative values by setting them to zero and linearly scales the sum of all attributions to one. We experimented with other normalization methods which keep the negative information, but ReLU works best. Appendix~\ref{sec:normalization_appendix} reports results
  for all normalization methods.}

Finally, we use bounding box information to aggregate the per-patch normalized attributions into per-object \new{attributions}. This is achieved using a mapping of objects onto a rectangular array of image patches, annotated manually such that no or only a non-significant portion of the object lies outside. To capture attributions that extend beyond the objects' boundaries (cf. Figure \ref{fig:model_structure}), we add a margin of one patch in all directions to each bounding box.

\section{Experimental Setup}

\subsection{Data}
\label{sec:data}

We use the visual and language stimuli used in the seminal visual world study presented in \citet{altmann99}.\footnote{We thank the authors of that study for making their original visual stimuli available to us.} \new{This dataset consists of 18 images, all of which we use.} Each image comes with a pair of captions of the form "The \emph{subject} will \emph{verb} the \emph{target object}". Figure \ref{fig:heatmaps} shows one such image. Each image shows five referents: one \emph{subject} performing the action and four \emph{object}s possible themes. The two actions (\emph{verb}s) were chosen:
\begin{compactitem}
    \item so that one of the verbs, called the \emph{constraining} verb, restricts the domain of subsequent reference to one object out of the four, which is called the \emph{target object} (cf. Figure \ref{fig:heatmaps}(b));
    \item and the other verb does not impose any such restriction and can feasibly combine with any of the four objects  (cf. Figure \ref{fig:heatmaps}(c)).
\end{compactitem}
%
%This idea is illustrated in Figure \ref{fig:heatmaps}, with the corresponding pair of captions being "The boy will eat the cake" and "The boy will move the cake".

\subsection{\new{Computing Gaze Preferences for Stimuli}}
\label{sec:models}

\new{A stimulus of a visual world experiment is a pair of an image and a language input (presented spoken in the experiment).} %Integrated Jacobians enable the user to ’select’ a set of language tokens and
%compute the attributions of these tokens to the image. 
We  compute attributions for each token position (cf. Figure \ref{fig:heatmaps}) as follows: (a) encode the pair of image and complete linguistic input once; (b) run Integrated Jacobians;
(c) compute sums over attribution scores for each prefix of the input (“the”, “the man”, etc.) to obtain the image attributions for this prefix. This procedure is only possible because the standard CLIP text encoders use masking to restrict attention to the left-hand context when computing the representation of a token \citep{radford2019language}. If it paid attention to the right-hand context as well, this would give
the encoder the opportunity to look 'into the future', jeopardizing the point of our experiment. That being said, our CLIP encoders are still genuine encoders trained on complete image-caption pairs and not  with a generative language modeling objective.

We use an implementation of Integrated Jacobians\footnote{\texttt{https://github.com/lucasmllr/exCLIP}} which supports both  original OpenAI CLIP \citep{clip} encoders and open-source OpenCLIP \citep{cherti2023reproducible} encoders. 
We select four CLIP encoder models (see Table \ref{tab:models}). We include the original OpenAI Vit-B/16 CLIP model from 2021 as starting point and compare it against more modern models which are trained on more data (the 1.3B Datacomp XL and 2B DFN datasets) and have more parameters. As the last column demonstrates, this scaling corresponds to generally higher zero-shot accuracy on the ImageNet benchmark (among others); \new{in an exploratory analysis, we ask} whether better ImageNet performance corresponds to better ability to account for human visual world behavior. Appendix\ \ref{sec:models_appendix} provides details on runtime and the checkpoints we use.

\paragraph{Baselines.} \new{We report two baselines to investigate the impact of visual and textual information, respectively. The first one computes the visual saliency \citep{KRONER2020261} for each object in our images and uses it in place of the object's attribution. The second one uses only the text encoder of the OpenAI ViT-B/16 CLIP model in an approach similar to \citet{allopenna1998tracking}: In each image, we define a language label for each object; for the target, this is the word used in the linguistic input. At each region, we use the text encoder to compute embeddings for the linguistic input and for the labels and use the cosine similarity between each object's label embedding and the embedding of the linguistic input in place of the object's attribution.}

We normalize the predictions of both baselines just like the CLIP bi-encoder predictions.

\subsection{Statistical Analysis}
\label{sec:statistical_analysis}

\new{Like standard analyses of human visual world data \citep{Ito_Knoeferle_2023}, we model population-level gaze preferences, not fixations, which depend on many further factors \citep{magnuson2019fixations}.  We ask whether our approach \textit{qualitatively} replicates human predictive processing, using two research questions:} (1) Does the approach predict the target object before it is mentioned when the verb is constraining? and (2) Is this predictive advantage specific to the pre-object region\new{, i.e., is it much reduced} once the object is explicitly named?

\begin{table*}[tb!h]
\centering
\begin{tabular}{lrcrcrc}
\toprule
& \multicolumn{2}{c}{Pre-object region} & \multicolumn{2}{c}{Object noun region} & \multicolumn{2}{c}{Interaction} \\
\cmidrule(r){2-3} \cmidrule(lr){4-5} \cmidrule(l){6-7}
Model & $\beta$ \new{(SE)} & $p$ & $\beta$ \new{(SE)} & $p$ & $\beta$ \new{(SE)} & $p$ \\
\midrule
 Text encoder BL & \new{0.007 (0.004)} & \new{0.081} & \new{0.001 (0.004)} & \new{0.708} & \new{$-0.005$ (0.005)} & \new{0.325} \\
 Visual saliency BL & \multicolumn{6}{c}{\new{no verb-type effect possible (identical predictions for both verbs)}} \\
\midrule
1 OpenAI (1) & \textbf{0.262} \new{(0.100)} & \textbf{0.011} & $-0.007$ \new{(0.100)} & 0.946 & $-0.269$ \new{(0.141)} & 0.062 \\
2 OpenCLIP (2) & \textbf{0.227} \new{(0.081)} & \textbf{0.007} & 0.055 \new{(0.081)} & 0.497 & $-0.171$ \new{(0.115)} & 0.141 \\
3 OpenCLIP (3) & \textbf{0.263} \new{(0.098)} & \textbf{0.010} & 0.115 \new{(0.098)} & 0.245 & $-0.147$ \new{(0.139)} & 0.292 \\
4 OpenCLIP (4) & \textbf{0.318} \new{(0.098)} & \textbf{0.002} & $-0.034$ \new{(0.098)} & 0.732 & \textbf{$-0.352$} \new{(0.138)} & \textbf{0.014} \\
\bottomrule
\end{tabular}
\caption{Linear mixed-effects model results for the four  CLIP models \new{and the baselines}. The table reports the estimated simple main effect of verb type (constraining minus non-constraining) within the pre-object region and within the object noun region, alongside the verb type $\times$ region interaction term\new{, with standard errors in parentheses}. The interaction term evaluates whether the constraining-verb advantage attenuates from the pre-object to the object noun region. Estimates and $p$-values are bolded where the effect is significant at $\alpha = 0.05$. Models numbered as in Table~1. \new{BL: baseline; text-encoder estimates are on a compressed scale (see Section~\ref{sec:results}).}}
\label{tab:per_model_results}
\end{table*}

We mirror the typical analysis of human data by analyzing the normalized attributions on the target object relative to the rest of the scene. Our dependent variable is  \textit{target preference}: the attribution to the target object minus mean attribution to all other objects. 
%We use the ReLU normalization (§\ref{sec:normalization}), which robustly bounds negative attributions.
Following the design of \citet{altmann99}, we analyze target preferences across two critical sentence regions: the pre-object prediction region (the verb and the subsequent article, e.g., ``eat the'') and the object noun region (e.g., ``cake''). As described in §\ref{sec:data},  we distinguish between \emph{constraining} verbs (``eat''), whose selectional restrictions single out the target object in the scene, and \emph{non-constraining} verbs (``move''), whose restrictions do not. The core predictive processing hypothesis asserts that a constraining verb drives a stronger preference for the target object in the pre-object region than a non-constraining verb, but that this difference attenuates at the target object.

We evaluate these claims using a two-level statistical approach.  First, to assess the consistency of the effect across CLIP models, we fit separate linear mixed-effects models for each of the four CLIP models.  Second, to test the predictive effect \new{across models}, we fit a global linear mixed-effects model wherein the four CLIP models are treated as a sampled random factor, analogous to human participants in an experiment\new{; with four models, generalization beyond these checkpoints is tentative}.  In both levels of analysis, the models predict target preference and include fixed effects for verb type, region, and their interaction.  Verb type and region are both entered using treatment contrast coding.  To evaluate our specific  claims, we extract the simple main effect of verb type (constraining minus non-constraining) within the pre-object region, the simple main effect of verb type within the object noun region, and their interaction.  The simple effects are obtained by rotating the reference level of the region factor between fits.
The per-model fits include random intercepts for experimental items (sentences).  The global model includes crossed random intercepts for items and CLIP models.  This random effect structure reflects the maximal design supported by the data that consistently converged across fits \citep{BARR2013255}. 

In sum, this approach tests two key claims corresponding to our research questions: (1) target preference is stronger for constraining verbs at the pre-object region (a positive simple effect of verb type at the pre-object reference level); and (2) this advantage is localized at the verb and diminishes at the object noun (a negative verb type $\times$ region interaction, \new{with the  effect of verb type at the object noun reference level reduced to close to zero}).

\section{Results}
\label{sec:results}

\begin{figure}[tb!]
    \centering
    \includegraphics[width=\linewidth]{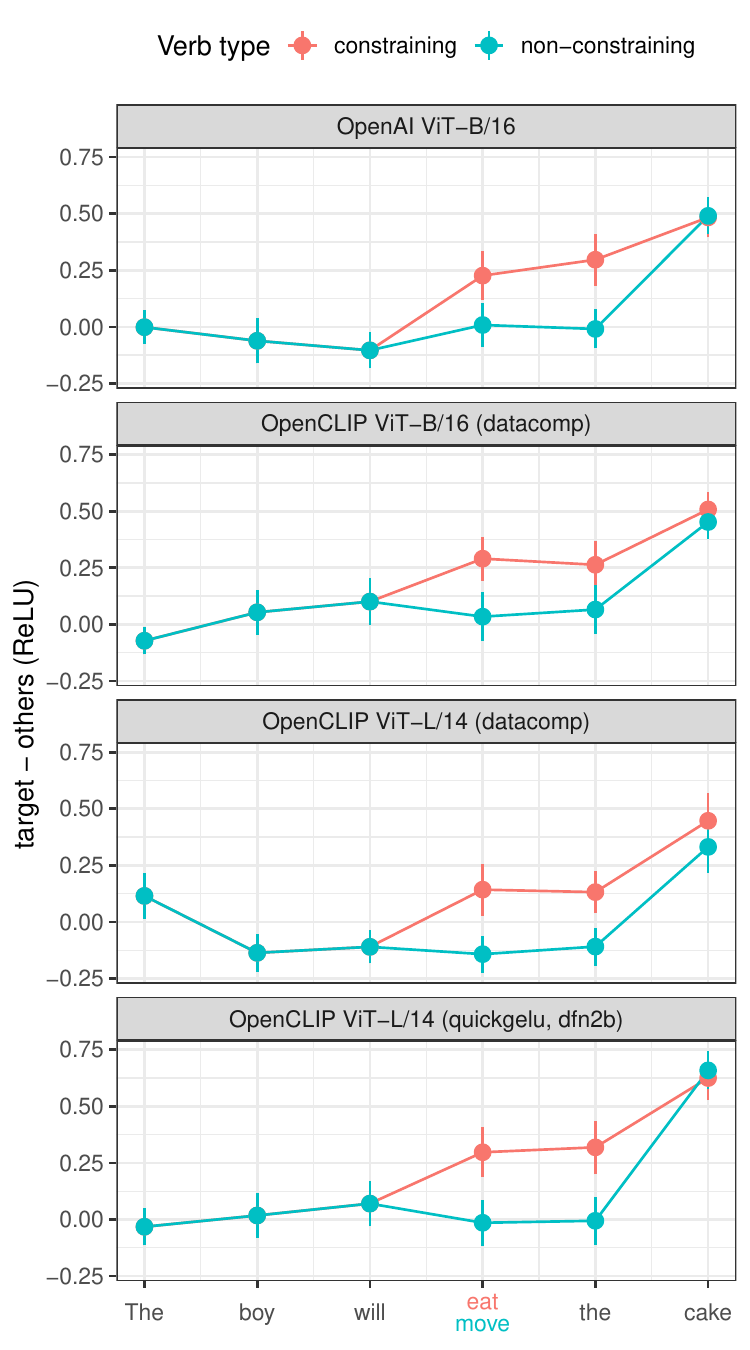}
    \caption{Target preference (attribution to target object minus mean attribution to other objects) by sentence position and verb type for each of the four tested CLIP models, utilizing ReLU normalization. For both verb types, preference for the target is initially low. Upon encountering a constraining verb (``eat''), target preference rises sharply prior to the target being named\new{, mirroring anticipatory looks in humans}. This predictive advantage over the non-constraining verb (``move'') \new{is much reduced} when the object is  named in the final position.\label{fig:target_pref} \vspace*{-1em}}
\end{figure}

\paragraph{Model-level results.}
Figure \ref{fig:target_pref} shows the target preference by sentence position and verb type for each of the four models, and Table \ref{tab:per_model_results} presents the results of the individual linear mixed-effects models fitted for each of them.\footnote{We evaluated a fifth checkpoint, OpenCLIP ViT-B/16 (dfn2b), but excluded it from analysis because its attributions are degenerate: inspection of its attribution maps revealed \emph{negative} attribution to the target object already at verb onset, so the relative target-preference measure is not interpretable for this model. We regard this as a model- or pipeline-specific artifact and leave its investigation to future work.}  In the pre-object region, all four models exhibit a significant constraining-verb advantage ($\beta \ge 0.227$, $p \le 0.011$), confirming that a constraining verb induces a stronger preference for the target object before it is explicitly named.  In the object noun region, as expected, none of the models exhibit a significant difference in target preference between the two verb types (all $p \ge 0.245$).  The interaction term, which explicitly tests whether the constraining-verb advantage attenuates from the pre-object region to the object noun, is negative for every model.  This attenuation reaches statistical significance for the OpenCLIP ViT-L/14 (quickgelu dfn2b) model ($p = 0.014$) and approaches significance for the OpenAI ViT-B/16 model ($p = 0.062$).  Given the limited statistical power of these per-model tests (which rely on only 18 experimental items), we next evaluate these claims \new{jointly} across all models.

\paragraph{\new{Across-model results.}}
\new{The} global mixed-effects model synthesizes the data across all four CLIP models, gaining substantial statistical power. In the critical pre-object region, we observe a strong and highly significant simple main effect of verb type ($\beta = 0.267$, $t = 4.60$, $p < 0.001$), confirming our primary claim: \new{across the four models}, the constraining verb induces predictive attribution to the target object before the object is encountered. In contrast, at the object noun region, the simple main effect of verb type is \new{close to zero} and non-significant ($\beta = 0.033$, $t = 0.56$, $p = 0.576$), indicating that the predictive advantage \new{is much reduced} once the visual object is explicitly named. Crucially, the interaction between verb type and region is significant and negative ($\beta = -0.235$, $t = -2.86$, $p = 0.005$), confirming that the constraining-verb advantage is localized to the predictive (pre-object) window. \new{Thus, for these materials, the four CLIP models qualitatively reproduce the anticipatory gaze pattern reported by \citet{altmann99}, despite not being trained for incremental language processing.}

\paragraph{\new{Baselines.}} \new{The visual saliency baseline cannot show a verb-type effect, since saliency does not depend on the linguistic input.  The text-only baseline shows a pre-object trend in the same direction as the CLIP models, but not significant ($t = 1.78$, $p = 0.081$; CLIP models: $t \ge 2.63$; Table~\ref{tab:per_model_results}).  Its estimates are on a compressed scale because the cosine similarities between the input and the four object labels differ only slightly; we therefore compare test statistics rather than estimates.  See Appendix~\ref{sec:baselines_appendix} for the time course.}

\paragraph{\new{Items without anticipation.}} Our best CLIP model \new{shows no} pre-object target preference notably for two stimuli with a constraining verb: \textit{The woman will bathe \dots} (target: \textit{baby}) and \textit{The chef will slice \dots} (target: \textit{fish}).  \new{The first is arguably not a model failure: without an object, \textit{bathe} is read reflexively \citep{ChristiansonEtAl2001}.  In the second case, the model fails to identify the fish throughout}; see Appendix~\ref{sec:failures-appendix} for details.

\paragraph{\new{Exploratory analysis: model quality.}} \new{In an exploratory analysis, we asked whether} the size of the pre-object constraining-verb effect (Table~\ref{tab:per_model_results}) tracks the model's ImageNet zero-shot accuracy (Table~\ref{tab:models}). The two OpenCLIP ViT-L/14 models, which score highest on ImageNet, also show the largest pre-object effects, while  OpenCLIP ViT-B/16  is lower on both counts \new{(Spearman $\rho = 0.8$)}. \new{The exception is}  OpenAI ViT-B/16, \new{which} shows a large pre-object effect despite \new{the lowest ImageNet accuracy}, possibly due to differences between open-source and proprietary models \citep{moeller2025explainingvisionlanguagesimilaritiesdual}.
\new{This pattern suggests a relation between a model's vision performance and the strength of its predictive effect, but it is not conclusive: with four models, even perfect rank agreement would not be significant (two-tailed $p = 0.083$), and ImageNet accuracy is confounded with model size.}

\section{Conclusions}
\label{sec:conclusions}

\new{In this paper, we asked whether standard bi-encoder vision-language models of the CLIP family can account for the time-course of human predictive processing in the visual-world paradigm. We presented an approach based on an 
attribution-based measure of target preference that  is  modest in its needs: We only require (a) pre-trained CLIP models; (b) annotations of bounding boxes (but not labels)
for the objects; and (c) model selection among
image encoders and normalization methods.}

We answer the research question affirmatively for the English experimental materials of \citet{altmann99} and these four models. \new{In all four CLIP models}, a constraining verb induced a \new{significantly} stronger preference for the target object already in the pre-object region, before the object was named, and this advantage \new{was much reduced} once the object was encountered. The key qualitative result is the interaction between verb type and region, %($\beta = -0.235$, $p = 0.005$), 
which establishes that the constraining-verb advantage is functionally localized to the predictive window rather than reflecting a global, region-independent bias. In this respect, our approach mirror the \new{anticipatory-gaze} logic of \citet{altmann99}: constraining lexical information is exploited before the referent becomes available, but this preference is time-locked to the point at which it is informative.

%($\beta = 0.267$, $p < 0.001$),d ($\beta = 0.033$, $p = 0.576$).

We regard the architecture of the underlying vision-language model as central.  Much recent work draws its cognitive models from autoregressive LMs whose training objective \emph{is} next-word prediction; when such a model reproduces human predictive processing, it is only exhibiting a behavior it was  optimized for.
%, and the explanatory gain is correspondingly modest.  
The bi-encoder CLIP models considered here carry no temporal or autoregressive objective: they are trained to align static images with static captions. That anticipatory processing nonetheless emerges from their attributions cannot be attributed to a training signal. This is therefore arguably a more informative finding.
%here than it would be for a model explicitly built to predict.

\new{An important caveat concerns the kind of prediction that our results demonstrate.  In the design of \citet{altmann99}, only the verb is manipulated.  The increased attribution to the cake after ``eat'' may therefore reflect a static association between verb and object, learned from co-occurrences in the pre-training data, rather than a prediction computed from the sentence as a whole.  Our design cannot distinguish these possibilities, but the same holds for the human data from this experiment.  Moreover, we do not regard these as competing explanations: the plausibility of verb-argument combinations is a major source of human prediction \citep{KamideEtAl2003}, so verb-object associations may be the basis of prediction rather than an alternative to it.  Accordingly, we do not claim that the bimodal architecture itself generates the prediction, but that it provides a linking function from the linguistic input to preferences over the objects in the scene.  However, note that the text encoder alone shows only a weak, non-significant trend in the same direction (Section~\ref{sec:results}), suggesting that verb information is present in the linguistic representation but that the effect becomes clear only when linked to the scene.  A design that separates association from prediction would contrast two plausible objects that differ in their association with the verb, e.g., two edible objects, one strongly and one weakly associated with ``eat'', or vary the agent while keeping the verb constant, as in \citet{KamideEtAl2003}.  This is future work.}

Finally, our approach assumes that higher attribution is a meaningful analogue of anticipatory human \new{gaze preferences}.  The precise relationship between \new{gaze preferences} and underlying processing is itself debated \citep[see, e.g.,][]{degen2021seeing}, but for the simple, static displays and token-level predictions considered here this linking assumption is a reasonable  approximation.

%Why does our approach succeed, then? \citet{moeller2025explainingvisionlanguagesimilaritiesdual} established that CLIP models identify object categories in images without being trained to do so, drawing from the correlational structure of the learned representation, since this helps the model distinguish matching from mismatching captions. Our results go one step further: models also acquire information about likely verb-object combinations. A plausible motivation is that events tend to be less imageable \citep{zhou2025quantifying} and therefore the presence (or absence) of objects in the image can provide important clues to the model.

We conclude from our study that cognitively plausible models of human language processing are not restricted to the generative (decoder) paradigm: Even very simple models can \new{qualitatively reproduce a classic anticipatory-gaze effect.  We speculate that anticipatory gaze} in the visual world paradigm may emerge from a similar multimodal similarity structure to the one that CLIP models acquire, without dedicated predictive machinery. We believe that this is a promising starting point for a discussion on the role of data vs. architecture in multimodal settings.

Our study opens future work in various directions. First, our present study rests on four models and one experiment. We need to look at more models to assess the generalization of our findings to the general class of CLIP models and gain enough statistical power to \new{test the relation between model quality and the predictive effect suggested by our exploratory analysis (Section~\ref{sec:results}).} Similarly, a broader range of experiments is needed to establish whether our approach can account for effects beyond predicate-argument prediction such as semantic competition \citep{Huettig2005WordMA} or color knowledge \citep{Huettig2011LookingAA}.
\new{A final direction is an exploration of the attention patterns in the text encoder to gain further insight into the predictive processes. We did not pursue this here given the concerns about the status of self-attention patterns \citep{jain-wallace-2019-attention}.}

% Our attributions, computed on prefixes, nonetheless recover the human
% time course. This suggests that the predictive signal is carried by
% the constraining verb rather than by incremental processing
% dynamics. Again, this analysis bears investigation on additional
% experiments.

%Beyond this main finding, we observed an exploratory trend by which the size of the pre-object effect tracks a model's ImageNet zero-shot accuracy. Whether stronger vision-language models systematically yield more human-like predictive behavior is a question for future work with a larger and more systematically varied set of checkpoints.

\section*{Limitations}

The main limitations of our approach are discussed in Section \ref{sec:conclusions}: First, our current study is limited to four CLIP models, and a single visual world experiment. We find encouraging results within this space, but more work is clearly necessary to delineate the exact scope of our claim regarding model classes and range of phenomena. \new{The sample is also small, with 18 items per model.  The per-model tests therefore have limited power, in particular for the verb type $\times$ region interaction, which reaches significance in only one of the four models.  Our main conclusions rest on the \new{across-model} analysis.}

\new{Second, our design manipulates only the verb and therefore cannot distinguish prediction based on the sentence as a whole from a static association between verb and object (see Section~\ref{sec:conclusions}).  Moreover, although the CLIP text encoder attends only to the left context, it is not trained on an incremental task, so the token-by-token time course that we derive from it is not a model of incremental processing in its own right.}

Third, our model is located at the granularity of tokens and cannot account for finer-grained effects.

\bibliography{custom}

@article{radford2019language,
  author    = {Alec Radford and Jeffrey Wu and Rewon Child and David Luan and Dario Amodei and Ilya Sutskever},
  title     = {Language Models are Unsupervised Multitask Learners},
  year      = {2019},
  journal   = {OpenAI Blog},
  volume    = {1},
  number    = {8}
}

@inproceedings{jain-wallace-2019-attention,
    title = "{A}ttention is not {E}xplanation",
    author = "Jain, Sarthak  and
      Wallace, Byron C.",
    editor = "Burstein, Jill  and
      Doran, Christy  and
      Solorio, Thamar",
    booktitle = "Proceedings of the 2019 Conference of the North {A}merican Chapter of the Association for Computational Linguistics: Human Language Technologies, Volume 1 (Long and Short Papers)",
    month = jun,
    year = "2019",
    address = "Minneapolis, Minnesota",
    publisher = "Association for Computational Linguistics",
    url = "https://aclanthology.org/N19-1357/",
    doi = "10.18653/v1/N19-1357",
    pages = "3543--3556"
}

@article{KRONER2020261,
title = {Contextual encoder–decoder network for visual saliency prediction},
journal = {Neural Networks},
volume = {129},
pages = {261-270},
year = {2020},
issn = {0893-6080},
doi = {https://doi.org/10.1016/j.neunet.2020.05.004},
url = {https://www.sciencedirect.com/science/article/pii/S0893608020301660},
author = {Alexander Kroner and Mario Senden and Kurt Driessens and Rainer Goebel}}

@Article{altmann99,
  author    = {Gerry T. M. Altmann and Yuki Kamide},
  title     = {Incremental interpretation at verbs: {Restricting} the domain of subsequent reference},
  journal   = {Cognition},
  url = {https://doi.org/10.1016/s0010-0277(99)00059-1},
  year      = 1999,
  volume    = 73,
  pages     = {247--264},
}

@Article{moeller2025explainingvisionlanguagesimilaritiesdual,
  author    = {Möller, Lucas and Tilli, Pascal and Vu, Ngoc Thang and Padó, Sebastian},
  title     = {Explaining Caption-Image Interactions in {CLIP} models with Second-Order Attributions},
  journal   = {Transactions on Machine Learning Research},
  year      = 2025,
  archiveprefix= {arXiv},
  eprint    = {2408.14153},
  primaryclass= {cs.CV},
  url       = {https://openreview.net/forum?id=HUUL19U7HP},
}

@InProceedings{ig,
  author    = {Mukund Sundararajan and Ankur Taly and Qiqi Yan},
  title     = {Axiomatic Attribution for Deep Networks},
  booktitle = {Proceedings of the 34th International Conference on Machine Learning},
  year      = {2017},
  editor    = {Precup, Doina and Teh, Yee Whye},
  volume    = {70},
  series    = {Proceedings of Machine Learning Research},
  pages     = {3319--3328},
  month     = {06--11 Aug},
  publisher = {PMLR},
  url       = {https://proceedings.mlr.press/v70/sundararajan17a.html},
}

@InProceedings{moeller-etal-2024-approximate,
  author    = {Möller, Lucas and Nikolaev, Dmitry and Pad{\'o}, Sebastian},
  title     = {Approximate Attributions for Off-the-Shelf {S}iamese Transformers},
  booktitle = {Proceedings of the 18th Conference of the European Chapter of the Association for Computational Linguistics},
  year      = {2024},
  editor    = {Graham, Yvette and Purver, Matthew},
  pages     = {2059--2071},
  month     = mar,
  address   = {St. Julian{'}s, Malta},
  publisher = {Association for Computational Linguistics},
  doi       = {10.18653/v1/2024.eacl-long.125},
  url       = {https://aclanthology.org/2024.eacl-long.125/},
}

@InProceedings{reimers-gurevych-2019-sentence,
  author    = {Reimers, Nils and Gurevych, Iryna},
  title     = {Sentence-{BERT}: {Sentence} Embeddings using {S}iamese {BERT}-Networks},
  booktitle = {Proceedings of the 2019 Conference on Empirical Methods in Natural Language Processing and the 9th International Joint Conference on Natural Language Processing (EMNLP-IJCNLP)},
  year      = {2019},
  editor    = {Inui, Kentaro and Jiang, Jing and Ng, Vincent and Wan, Xiaojun},
  pages     = {3982--3992},
  month     = {nov},
  address   = {Hong Kong, China},
  publisher = {Association for Computational Linguistics},
  url       = {https://aclanthology.org/D19-1410},
  doi       = {10.18653/v1/D19-1410},
}

@Article{OhSchuler2023,
  author    = {Oh, Byung-Doh and Schuler, William},
  title     = {Why Does Surprisal From Larger Transformer-Based Language Models Provide a Poorer Fit to Human Reading Times?},
  journal   = {Transactions of the Association for Computational Linguistics},
  year      = {2023},
  volume    = {11},
  pages     = {336-350},
  doi       = {10.1162/tacl_a_00548},
}

@Article{CuskleyEtAl2024,
  author    = {Cuskley, Christine and Woods, Rebecca and Flaherty, Molly},
  title     = {The Limitations of Large Language Models for Understanding Human Language and Cognition},
  journal   = {Open Mind},
  year      = {2024},
  volume    = {8},
  pages     = {1058-1083},
  month     = {08},
  issn      = {2470-2986},
  doi       = {10.1162/opmi_a_00160},
}

@InProceedings{Wilcox2020OnTPA,
  author    = {Ethan Gotlieb Wilcox and Jon Gauthier and Jennifer Hu and Peng Qian and Roger Levy},
  title     = {On the Predictive Power of Neural Language Models for Human Real-Time Comprehension Behavior},
  booktitle = {Proceedings of CogSci},
  year      = {2020},
}

@Article{VanSchijndelLinzen2021,
  author    = {van Schijndel, Marten and Linzen, Tal},
  title     = {Single-Stage Prediction Models Do Not Explain the Magnitude of Syntactic Disambiguation Difficulty},
  journal   = {Cognitive Science},
  year      = {2021},
  volume    = {45},
  number    = {6},
  month     = {6},
  pages = {e12988},
  issn      = {1551-6709},
  doi       = {10.1111/cogs.12988},
  publisher = {Wiley},
}

@InProceedings{frank-2024-neural,
    title = "Neural language model gradients predict event-related brain potentials",
    author = "Frank, Stefan L.",
    editor = "Futrell, Richard  and
      Mayer, Connor  and
      Zaslavsky, Noga",
    booktitle = "Proceedings of the Society for Computation in Linguistics 2024",
    month = jun,
    year = "2024",
    address = "Irvine, CA",
    publisher = "Association for Computational Linguistics",
    url = "https://aclanthology.org/2024.scil-1.24/",
    pages = "316--323"
}

@article{Huettig2011LookingAA,
  author    = {Falk Huettig and Gerry T. M. Altmann},
  title     = {Looking at anything that is green when hearing ``frog'': how object surface colour and stored object colour knowledge influence language-mediated overt attention},
  journal   = {Quarterly Journal of Experimental Psychology},
  volume    = {64},
  number    = {1},
  pages     = {122--145},
  year      = {2011},
  doi       = {10.1080/17470218.2010.481474}
}

@article{Huettig2005WordMA,
  author    = {Falk Huettig and Gerry T. M. Altmann},
  title     = {Word meaning and the control of eye fixation: semantic competitor effects and the visual world paradigm},
  journal   = {Cognition},
  volume    = {96},
  number    = {1},
  pages     = {B23--B32},
  year      = {2005},
  doi       = {10.1016/j.cognition.2004.10.003}
}

@inproceedings{shen2021clipbenefitvisionandlanguagetasks,
      title={How Much Can {CLIP} Benefit Vision-and-Language Tasks?}, 
      author={Sheng Shen and Liunian Harold Li and Hao Tan and Mohit Bansal and Anna Rohrbach and Kai-Wei Chang and Zhewei Yao and Kurt Keutzer},
      year={2021},
      booktitle = {Proceedings of ICLR},
      url={https://arxiv.org/abs/2107.06383}, 
}

@InProceedings{Toneva2019InterpretingAIA,
  author    = {Mariya Toneva and Leila Wehbe},
  title     = {Interpreting and improving natural-language processing (in machines) with natural language-processing (in the brain)},
  booktitle = {Proceedings of NeurIPS},
  year      = {2019},
}

@Article{LinzenBaroni2021,
  author    = {Linzen, Tal and Baroni, Marco},
  title     = {Syntactic Structure from Deep Learning},
  journal   = {Annual Review of Linguistics},
  year      = {2021},
  volume    = {7},
  number    = {1},
  pages     = {195--212},
  month     = {1},
  day       = {14},
  publisher = {Annual Reviews},
  issn      = {2333-9683},
  doi       = {10.1146/annurev-linguistics-032020-051035},
}

@Article{https://doi.org/10.1111/lnc3.70001,
  author    = {Portelance, Eva and Jasbi, Masoud},
  title     = {The Roles of Neural Networks in Language Acquisition},
  journal   = {Language and Linguistics Compass},
  year      = {2024},
  volume    = {18},
  number    = {6},
  pages     = {e70001},
  doi       = {https://doi.org/10.1111/lnc3.70001},
  url       = {https://compass.onlinelibrary.wiley.com/doi/abs/10.1111/lnc3.70001},
  eprint    = {https://compass.onlinelibrary.wiley.com/doi/pdf/10.1111/lnc3.70001},
}

@InProceedings{degen2021seeing,
  author    = {Degen, Judith and Kursat, Leyla and Leigh, Daisy Dorothy},
  title     = {Seeing is believing: Testing an explicit linking assumption for visual world eye-tracking in psycholinguistics},
  booktitle = {Proceedings of the Annual Meeting of the Cognitive Science Society},
  year      = {2021},
  url       = {https://escholarship.org/uc/item/6182t9jb},
}

@Article{cooper1974control,
  author    = {Cooper, Roger M},
  title     = {The control of eye fixation by the meaning of spoken language: a new methodology for the real-time investigation of speech perception, memory, and language processing.},
  journal   = {Cognitive psychology},
  volume = "6",
  number = "1",
  pages = "84-107",
  year      = {1974},
}

@Article{yoon2018influence,
  author    = {Yoon, Si On and Brown-Schmidt, Sarah},
  title     = {Influence of the historical discourse record on language processing in dialogue},
  journal   = {Discourse Processes},
  year      = {2018},
  volume    = {55},
  number    = {1},
  pages     = {31--46},
  publisher = {Routledge},
}

@InProceedings{medclip,
  author    = {Zhang, Yuhao and Jiang, Hang and Miura, Yasuhide and Manning, Christopher D and Langlotz, Curtis P},
  title     = {Contrastive learning of medical visual representations from paired images and text},
  booktitle = {Machine Learning for Healthcare Conference},
  year      = {2022},
}

@Article{wolfe2017five,
  author    = {Wolfe, Jeremy M. and Horowitz, Todd S.},
  title     = {Five Factors that Guide Attention in Visual Search},
  journal   = {Nature Human Behaviour},
  year      = {2017},
  volume    = {1},
  pages     = {0058},
  month     = {Mar},
  doi       = {10.1038/s41562-017-0058},
}

@Article{
    murdoch,
  author    = {W. James Murdoch and Chandan Singh and Karl Kumbier and Reza Abbasi-Asl and Bin Yu},
  title     = {Definitions, methods, and applications in interpretable machine learning},
  url = {https://doi.org/10.1073/pnas.1900654116},
  journal   = {Proceedings of the National Academy of Sciences},
  year      = {2019},
}

@article{BARR2013255,
title = {Random effects structure for confirmatory hypothesis testing: Keep it maximal},
journal = {Journal of Memory and Language},
volume = {68},
number = {3},
pages = {255-278},
year = {2013},
issn = {0749-596X},
doi = {https://doi.org/10.1016/j.jml.2012.11.001},
url = {https://www.sciencedirect.com/science/article/pii/S0749596X12001180},
author = {Dale J. Barr and Roger Levy and Christoph Scheepers and Harry J. Tily}}

@incollection{Rumelhart1986PastTense,
  author    = {Rumelhart, David E. and McClelland, James L.},
  title     = {On learning the past tense of {E}nglish verbs},
  booktitle = {Parallel Distributed Processing: Explorations in the Microstructure of Cognition, Volume 2: Psychological and Biological Models},
  editor    = {McClelland, James L. and Rumelhart, David E. and the PDP Research Group},
  pages     = {216--271},
  publisher = {MIT Press},
  address   = {Cambridge, MA},
  year      = {1986}
}

@article{allopenna1998tracking,
  author    = {Allopenna, Paul D. and Magnuson, James S. and Tanenhaus, Michael K.},
  title     = {Tracking the Time Course of Spoken Word Recognition Using Eye Movements: Evidence for Continuous Mapping Models},
  journal   = {Journal of Memory and Language},
  volume    = {38},
  number    = {4},
  pages     = {419--439},
  year      = {1998},
  doi       = {10.1006/jmla.1997.2558},
}

@article{mcclelland1986trace,
  author    = {McClelland, James L. and Elman, Jeffrey L.},
  title     = {The {TRACE} Model of Speech Perception},
  journal   = {Cognitive Psychology},
  volume    = {18},
  number    = {1},
  pages     = {1--86},
  year      = {1986},
  doi       = {10.1016/0010-0285(86)90015-0},
  pmid      = {3753912}
}

@inproceedings{Liu2023VisualInstruction,
  author    = {Liu, Haotian and Li, Chunyuan and Wu, Qingyang and Lee, Yong Jae},
  title     = {Visual Instruction Tuning},
  booktitle = {Proceedings of NeurIPS},
  year      = {2023}
}

@article{Bai2025Qwen25VL,
  author    = {Bai, Shuai and Chen, Keqin and Liu, Xuejing and Wang, Jialin and Ge, Wenbin and Song, Sibo and Dang, Kai and Wang, Peng and Wang, Shijie and Tang, Jun and Zhong, Humen and Zhu, Yuanzhi and Yang, Mingkun and Li, Zhaohai and Wan, Jianqiang and Wang, Pengfei and Ding, Wei and Fu, Zheren and Xu, Yiheng and Ye, Jiabo and Zhang, Xi and Xie, Tianbao and Cheng, Zesen and Zhang, Hang and Yang, Zhibo and Xu, Haiyang and Lin, Junyang},
  title     = {Qwen2.5-{VL} Technical Report},
  journal   = {arXiv preprint arXiv:2502.13923},
  year      = {2025}
}

@inproceedings{fang2023datafilteringnetworks,
      title={Data Filtering Networks}, 
      author={Alex Fang and Albin Madappally Jose and Amit Jain and Ludwig Schmidt and Alexander Toshev and Vaishaal Shankar},
      booktitle = {Proceedings of ICLR},
      year={2024},
      url={https://arxiv.org/abs/2309.17425}, 
}

@inproceedings{gadre2023datacompsearchgenerationmultimodal,
      title={DataComp: In search of the next generation of multimodal datasets}, 
      author={Samir Yitzhak Gadre and Gabriel Ilharco and Alex Fang and Jonathan Hayase and Georgios Smyrnis and Thao Nguyen and Ryan Marten and Mitchell Wortsman and Dhruba Ghosh and Jieyu Zhang and Eyal Orgad and Rahim Entezari and Giannis Daras and Sarah Pratt and Vivek Ramanujan and Yonatan Bitton and Kalyani Marathe and Stephen Mussmann and Richard Vencu and Mehdi Cherti and Ranjay Krishna and Pang Wei Koh and Olga Saukh and Alexander Ratner and Shuran Song and Hannaneh Hajishirzi and Ali Farhadi and Romain Beaumont and Sewoong Oh and Alex Dimakis and Jenia Jitsev and Yair Carmon and Vaishaal Shankar and Ludwig Schmidt},
      booktitle = {Proceedings of NeurIPS -- Datasets and Benchmarks Track},
      year={2023},
      eprint={2304.14108},
      archivePrefix={arXiv},
      primaryClass={cs.CV},
      url={https://arxiv.org/abs/2304.14108}, 
}

@InProceedings{clip,
  author    = {Radford, Alec and Kim, Jong Wook and Hallacy, Chris and Ramesh, Aditya and Goh, Gabriel and Agarwal, Sandhini and Sastry, Girish and Askell, Amanda and Mishkin, Pamela and Clark, Jack and Krueger, Gretchen and Sutskever, Ilya},
  title     = {Learning Transferable Visual Models From Natural Language Supervision},
  booktitle = {Proceedings of the 38th International Conference on Machine Learning},
  year      = {2021},
}

@Article{magnuson2019fixations,
  author    = {Magnuson, James S.},
  title     = {Fixations in the visual world paradigm: where, when, why?},
  journal   = {Journal of Cultural Cognitive Science},
  year      = {2019},
  volume    = {3},
  number    = {2},
  pages     = {113--139},
  doi       = {10.1007/s41809-019-00035-3},
  url       = {https://doi.org/10.1007/s41809-019-00035-3},
}

@Article{Ito_Knoeferle_2023,
  author    = {Ito, Aine and Knoeferle, Pia},
  title     = {Analysing data from the psycholinguistic visual-world paradigm: Comparison of different analysis methods},
  journal   = {Behavior Research Methods},
  year      = {2023},
  volume    = {55},
  number    = {7},
  pages     = {3461--3493},
  doi       = {10.3758/s13428-022-01969-3},
  url       = {https://link.springer.com/article/10.3758/s13428-022-01969-3},
}

@Article{huettig2011using,
  author    = {Huettig, Falk and Rommers, Joost and Meyer, Antje S},
  title     = {Using the visual world paradigm to study language processing: A review and critical evaluation},
  journal   = {Acta psychologica},
  year      = {2011},
  volume    = {137},
  number    = {2},
  url = {https://doi.org/10.1016/j.actpsy.2010.11.003},
  pages     = {151--171},
  publisher = {Elsevier},
}

@article{mirman2008statistical,
  author    = {Mirman, Daniel and Dixon, James A. and Magnuson, James S.},
  title     = {Statistical and computational models of the visual world paradigm: Growth curves and individual differences},
  journal   = {Journal of Memory and Language},
  volume    = {59},
  number    = {4},
  pages     = {475--494},
  year      = {2008},
  month     = {nov},
  doi       = {10.1016/j.jml.2007.11.006},
}

@article{BRUCE201595,
title = {On computational modeling of visual saliency: Examining what’s right, and what’s left},
journal = {Vision Research},
volume = {116},
pages = {95-112},
year = {2015},
issn = {0042-6989},
doi = {https://doi.org/10.1016/j.visres.2015.01.010},
url = {https://www.sciencedirect.com/science/article/pii/S0042698915000267},
author = {Neil D.B. Bruce and Calden Wloka and Nick Frosst and Shafin Rahman and John K. Tsotsos}}

@Article{tanenhaus1995,
  author    = {Tanenhaus, Michael K and Spivey-Knowlton, Michael J and Eberhard, Kathleen M and Sedivy, Julie C},
  title     = {Integration of visual and linguistic information in spoken language comprehension},
  journal   = {Science},
  url = {https://doi.org/10.1126/science.7777863},
  year      = {1995},
  volume    = {268},
  number    = {5217},
  pages     = {1632--1634},
  publisher = {American Association for the Advancement of Science},
}

@Article{sedivy1999,
  author    = {Sedivy, Julie C and Tanenhaus, Michael K and Chambers, Craig G and Carlson, Greg N},
  title     = {Achieving incremental semantic interpretation through contextual representation},
  url = {https://doi.org/10.1016/s0010-0277(99)00025-6},
  journal   = {Cognition},
  year      = {1999},
  volume    = {71},
  number    = {2},
  pages     = {109--147},
  publisher = {Elsevier},
}

@InProceedings{cherti2023reproducible,
  author    = {Cherti, Mehdi and Beaumont, Romain and Wightman, Ross and Wortsman, Mitchell and Ilharco, Gabriel and Gordon, Cade and Schuhmann, Christoph and Schmidt, Ludwig and Jitsev, Jenia},
  title     = {Reproducible scaling laws for contrastive language-image learning},
  booktitle = {Proceedings of the IEEE/CVF Conference on Computer Vision and Pattern Recognition (CVPR)},
  year      = {2023},
  pages     = {2818--2829},
}

@Article{KamideEtAl2003,
  author    = {Kamide, Yuki and Altmann, Gerry T.M and Haywood, Sarah L},
  title     = {The time-course of prediction in incremental sentence processing: {Evidence} from anticipatory eye movements},
  journal   = {Journal of Memory and Language},
  year      = {2003},
  volume    = {49},
  number    = {1},
  pages     = {133--156},
  month     = jul,
  issn      = {0749-596X},
  doi       = {10.1016/s0749-596x(03)00023-8},
  publisher = {Elsevier BV},
}

@Manual{RCoreTeam2024,
  title        = {R: A Language and Environment for Statistical Computing},
  author       = {{R Core Team}},
  organization = {R Foundation for Statistical Computing},
  address      = {Vienna, Austria},
  year         = {2024},
  url          = {https://www.R-project.org/},
}

@Article{BatesEtAl2015,
  author    = {Bates, Douglas and M{\"a}chler, Martin and Bolker, Ben and Walker, Steve},
  title     = {Fitting Linear Mixed-Effects Models Using {lme4}},
  journal   = {Journal of Statistical Software},
  year      = {2015},
  volume    = {67},
  number    = {1},
  pages     = {1--48},
  doi       = {10.18637/jss.v067.i01},
}

@Article{KuznetsovaEtAl2017,
  author    = {Kuznetsova, Alexandra and Brockhoff, Per B. and Christensen, Rune H. B.},
  title     = {{lmerTest} Package: Tests in Linear Mixed Effects Models},
  journal   = {Journal of Statistical Software},
  year      = {2017},
  volume    = {82},
  number    = {13},
  pages     = {1--26},
  doi       = {10.18637/jss.v082.i13},
}

@Article{ChristiansonEtAl2001,
  author    = {Kiel Christianson and Andrew Hollingworth and John F. Halliwell and Fernanda Ferreira},
  title     = {Thematic Roles Assigned along the Garden Path Linger},
  journal   = {Cognitive Psychology},
  year      = {2001},
  volume    = {42},
  number    = {4},
  pages     = {368--407},
  doi       = {10.1006/cogp.2001.0752},
  publisher = {Elsevier},
}

\appendix

\section{Modeling Details}
\label{sec:models_appendix}

\paragraph{CLIP Model URLs.} We used the following HuggingFace URLs to obtain the CLIP models used in the experiment.
\begin{itemize}
    \item OpenAI ViT-B/16: \url{https://huggingface.co/openai/clip-vit-base-patch16}
    \item OpenCLIP ViT-B/16 (datacomp xl): \url{https://huggingface.co/flavour/CLIP-ViT-B-16-DataComp.XL-s13B-b90K}{}
    \item OpenCLIP ViT-L/14 (datacomp xl): \url{https://huggingface.co/laion/CLIP-ViT-L-14-DataComp.XL-s13B-b90K}
    \item OpenCLIP ViT-L/14 (quickgelu dfn2b): \url{https://huggingface.co/apple/DFN2B-CLIP-ViT-L-14}
\end{itemize}

\paragraph{Runtime.} Our experiments were carried out on servers with Nvidia GeForce RTX A6000 and Nvidia RTX 6000 Ada GPUs with 48 GB RAM. Walltime for one complete run (encoding, attribution, aggregation) was between 2 minutes and  10 minutes depending on model size.

\paragraph{Regression Models.}

Regression models were fit in R 4.4.1 \citep{RCoreTeam2024} with lme4 1.1-37 \citep{BatesEtAl2015}; $p$-values were computed with lmerTest 3.2-1 \citep[Satterthwaite approximation;][]{KuznetsovaEtAl2017}.

\begin{table*}[bth]
\centering
\begin{tabular}{lrrrrrrr}
\toprule
& & \multicolumn{2}{c}{Pre-object region} & \multicolumn{2}{c}{Object noun region} & \multicolumn{2}{c}{Interaction} \\
\cline{3-4} \cline{5-6} \cline{7-8}
Model & Norm. & $\beta$ & $p$ & $\beta$ & $p$ & $\beta$ & $p$ \\
\midrule
1a OpenAI ViT-B/16 & None & 0.226 & 0.354 & 0.049 & 0.841 & $-0.178$ & 0.606 \\
1b OpenAI ViT-B/16 & ReLU & \textbf{0.262} & \textbf{0.011} & $-0.007$ & 0.946 & $-0.269$ & 0.062 \\
1c OpenAI ViT-B/16 & Shift & \textbf{0.21} & \textbf{0.006} & 0.014 & 0.85 & $-0.196$ & 0.063 \\
1d OpenAI ViT-B/16 & Softmax & 0.001 & 0.187 & 0 & 0.592 & 0 & 0.575 \\
\midrule
2a OpenCLIP ViT-B/16 & None & $-0.834$ & 0.363 & $-0.31$ & 0.734 & 0.524 & 0.685 \\
2b OpenCLIP ViT-B/16 & ReLU & \textbf{0.227} & \textbf{0.007} & 0.055 & 0.497 & $-0.171$ & 0.141 \\
2c OpenCLIP ViT-B/16 & Shift & 0.098 & 0.103 & 0.048 & 0.415 & $-0.049$ & 0.556 \\
2d OpenCLIP ViT-B/16 & Softmax & 0.001 & 0.111 & 0.001 & 0.165 & 0 & 0.88 \\
\midrule

3a OpenCLIP ViT-L/14  & None & -0.152 & 0.877 & 1.24 & 0.21 & 1.39 & 0.318 \\
3b OpenCLIP ViT-L/14  & ReLU & \textbf{0.263} & \textbf{0.010} & 0.115 & 0.245 & $-0.147$ & 0.292 \\
3c OpenCLIP ViT-L/14  & Shift & \textbf{0.22} & \textbf{$<$0.001} & 0.107 & 0.064 & $-0.113$ & 0.166 \\
3d OpenCLIP ViT-L/14  & Softmax & \textbf{0.001} &  \textbf{0.035} & 0.001 & 0.273 & $-0.001$ &  0.459 \\
\midrule

4a OpenCLIP ViT-L/14  & None & $-4.42$ & 0.199 & $-2.83$ & 0.409 & 1.59 & 0.742 \\
4b OpenCLIP ViT-L/14  & ReLU & \textbf{0.318} & \textbf{0.002} & $-0.034$ & 0.732 & \textbf{$-0.352$} & \textbf{0.014} \\
4c OpenCLIP ViT-L/14  & Shift & \textbf{0.256} & \textbf{$<$0.001} & 0.01 & 0.858 & \textbf{$-0.246$} & \textbf{0.003} \\
4d OpenCLIP ViT-L/14  & Softmax & \textbf{0.002} & \textbf{0.005} & 0 & 0.646 & $-0.002$ & 0.087 \\

\bottomrule
\end{tabular}
\caption{Results for the four CLIP models with four normalization methods. See Table \ref{tab:per_model_results} for interpretation details.}
\label{tab:per_model_results_normalization_appendix}
\end{table*}

\section{Details on Normalization Methods}
\label{sec:normalization_appendix}

We experimented with four normalization methods. Let $a_0(bb, l)$ be the raw attribution score between a linguistic input $l$ and all image patches inside a bounding box $bb$. Then we can define a normalized attribution score $a_x$ using a normalization strategy $x$ as follows:
\begin{description}
    \item[None]  $a_\text{None}(bb, l) = a_0(bb, l)$. This method does not normalize.
    \item[ReLU] $a_\text{ReLu}(bb, l) = \max(0,a_0(bb, l))$. This method clamps all negative attributions to zero.
    \item[Shift] $a_\text{Shift}(bb, l) = a_0(bb, l) + c$ where $c=\min_{bb'} a_0(bb', l)$. This method determines the bounding box with the smallest attribution and shifts it to zero. All other scores are shifted accordingly.
    \item[Softmax] $a_\text{Softmax}(bb, l) = \frac{\exp(a_0(bb, l))}{\sum_{bb'} \exp(a_0(bb', l))}$. This method applies the well-known softmax transformation to the scores.
\end{description}
The results are shown in Table~\ref{tab:per_model_results_normalization_appendix}. In short, they indicate that negative attribution scores are unreliable, or at least uninformative, for the task at hand: The \new{method} that simply throws away negative attribution scores (ReLU) performs best in all respects (consistent effect in pre-object region for all four models). The Shift method, which keeps the linear structure of the attribution scores intact, comes in second place with pre-object region effects for 3 out of 4. \new{It is the only normalization method other than ReLU which finds a significant interaction, but only for one model.} The Softmax method tends to greatly compress the predicted attribution scores (compare the very small $\beta$ values) and only manages to obtain a significant pre-object region effect for \new{two models}. Without normalization (None), we do not see any effects.

\new{The second observation is descriptive and in line with the exploratory analysis in Section~\ref{sec:results}: the two larger ViT-L/14 models show the pre-object effect under three normalization strategies (ReLU, Shift, Softmax), the two smaller models under at most two.  For larger models, the choice of normalization thus appears to matter less.} 

\section{\new{Baselines}}
\label{sec:baselines_appendix}

\begin{figure}[tb!]
    \centering
    \includegraphics[width=\linewidth]{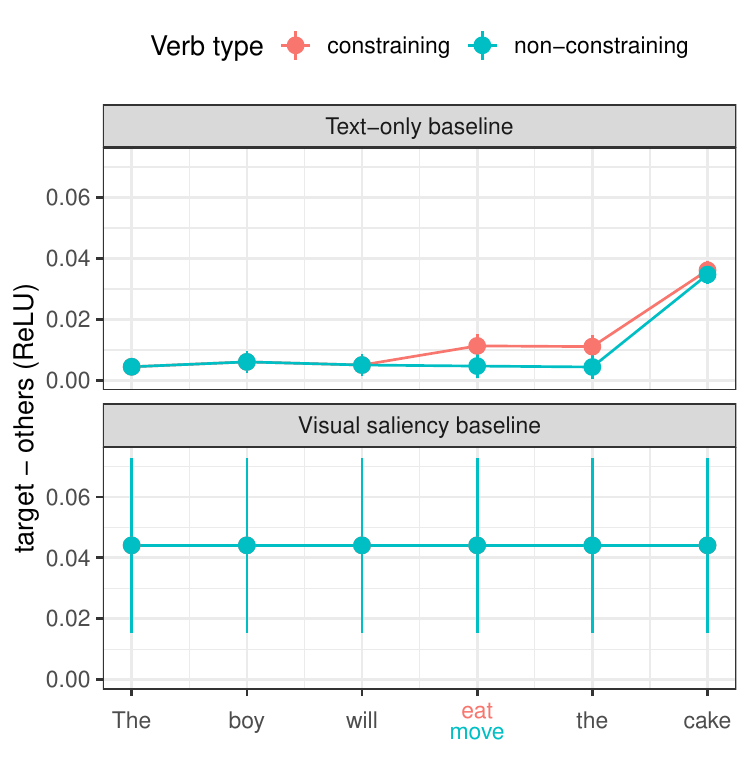}
    \caption{\new{Target preference by sentence position and verb type for the text-only and visual saliency baselines (ReLU normalization).  Note the different y-axis scale compared to Figure~\ref{fig:target_pref}.}}
    \label{fig:baselines}
\end{figure}

\new{Figure~\ref{fig:baselines} shows the time course of target preference for the two baselines.  The visual saliency baseline does not depend on the linguistic input, so its target preference is constant across positions and identical for both verb types.  The text-only baseline shows a small advantage for the constraining verb at the verb and the article, which is not significant (Table~\ref{tab:per_model_results}), and an increase in target preference for both verb types when the object is named.}

\section{\new{Heatmaps for Items Without Anticipation}}
\label{sec:failures-appendix}

\begin{figure*}[tb!h]
    \centering
    \begin{subfigure}[t]{0.3\linewidth}
    \centering
        \includegraphics[width=\linewidth]{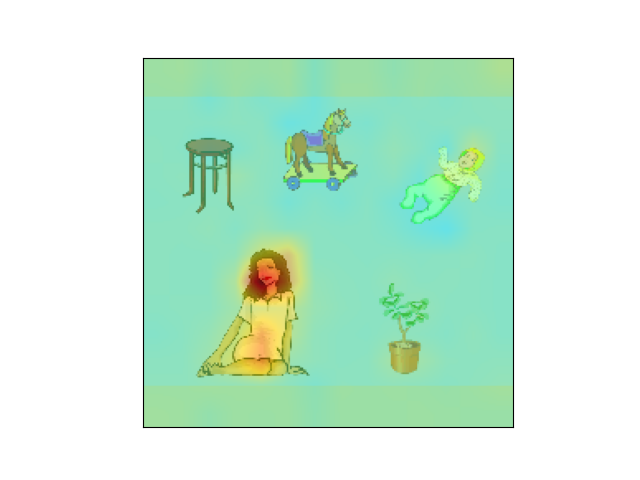}
        \subcaption{'the woman'}
    \end{subfigure}\hfill
    \begin{subfigure}[t]{0.3\linewidth}
        \centering
        \includegraphics[width=\linewidth]{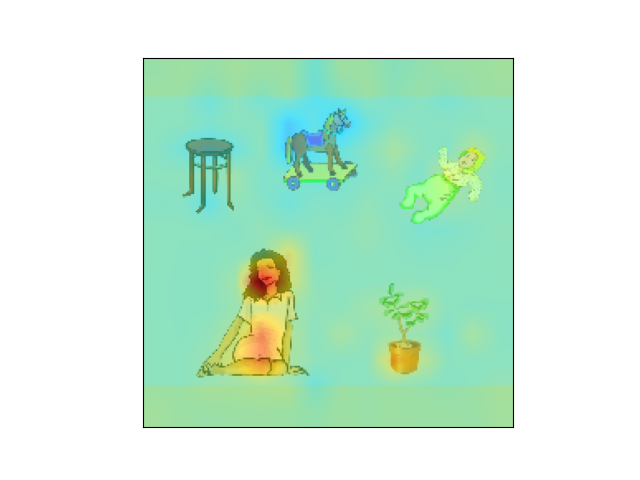}
        \subcaption{'the woman will bathe'}
    \end{subfigure}\hfill
    \begin{subfigure}[t]{0.3\linewidth}
        \centering
        \includegraphics[width=\linewidth]{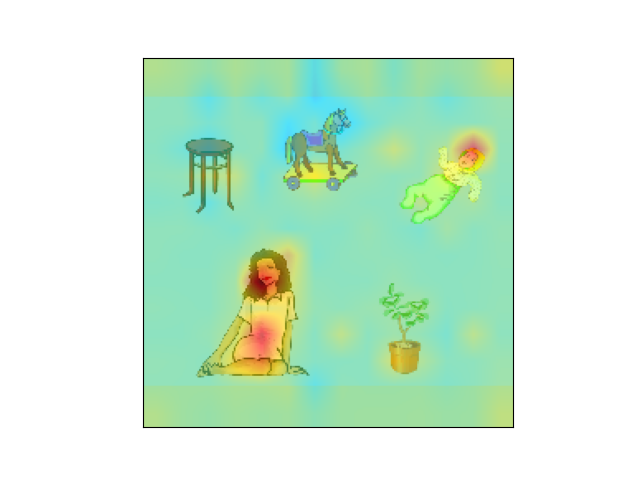}
        \subcaption{'the woman will bathe the baby'}
    \end{subfigure}
    \caption{Heatmaps for sentence 4 with constraining verb: \new{reflexive reading of \textit{bathe}}}
    \label{fig:heatmap-failure1}
\end{figure*}
\begin{figure*}[tb!]
    \centering
    \begin{subfigure}[t]{0.3\linewidth}
    \centering
        \includegraphics[width=\linewidth]{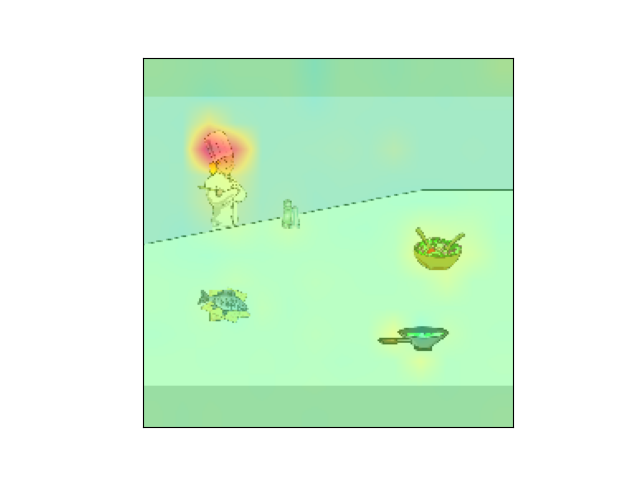}
        \subcaption{'the chef'}
    \end{subfigure}\hfill
    \begin{subfigure}[t]{0.3\linewidth}
        \centering
        \includegraphics[width=\linewidth]{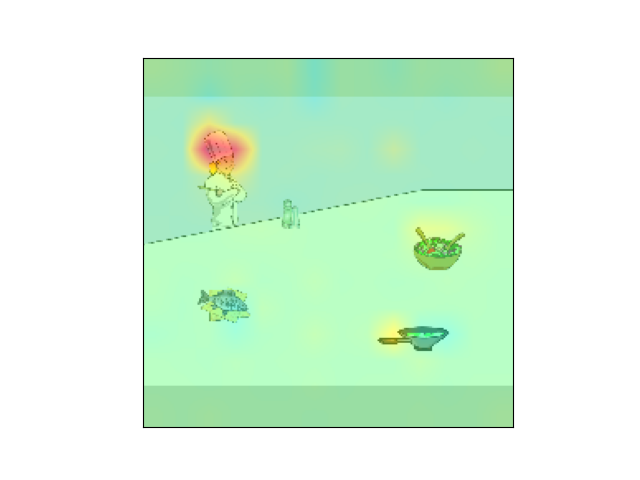}
        \subcaption{'the chef will slice'}
    \end{subfigure}\hfill
    \begin{subfigure}[t]{0.3\linewidth}
        \centering
        \includegraphics[width=\linewidth]{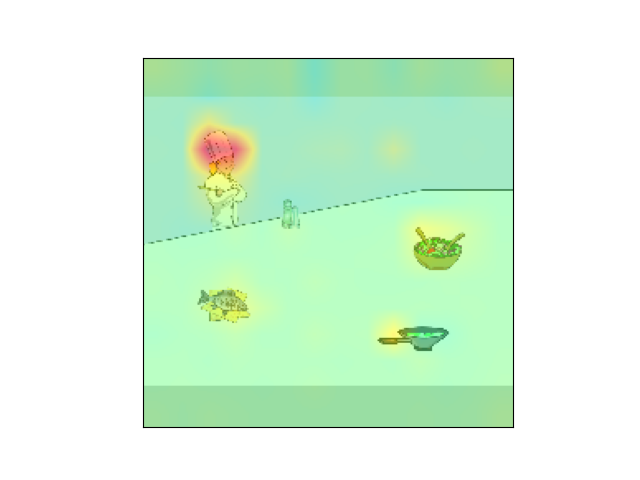}
        \subcaption{'the chef will slice the fish'}
    \end{subfigure}
    \caption{Heatmaps for sentence 18: Failure to identify target object}
    \label{fig:heatmap-failure2}
\end{figure*}

We show heatmaps for sentences 4 and 18 from the experimental materials, as produced by the overall best model, OpenCLIP ViT-L/14 (quickgelu dfn2b). The model \new{does not show} the expected target preferences in the pre-object region for the sentences in the constraining verb conditions. \new{For sentence~4, however, this is arguably not a failure of the model.  \textit{Bathe} is a reflexive absolute transitive verb: in the absence of a direct object, it is interpreted reflexively, i.e., \textit{The woman bathed} means that she bathed herself \citep{ChristiansonEtAl2001}.  At the verb, attribution to the woman rather than the baby (Figure~\ref{fig:heatmap-failure1}(b)) is therefore compatible with the preferred reading, and the model recognizes the baby when it is mentioned (Figure~\ref{fig:heatmap-failure1}(c)).  This item is thus not a clean test of verb-based anticipation.}  In sentence~18, \new{by contrast,} the model fails to form expectations at the verb (Figure~\ref{fig:heatmap-failure2}(b)) and does not even recognize the object when it is mentioned (Figure~\ref{fig:heatmap-failure2}(c)).

\end{document}